\documentclass[letterpaper, 10 pt, conference]{ieeeconf}  % Comment this line out if you need a4paper

\IEEEoverridecommandlockouts                              % This command is only needed if 
\usepackage{graphics} % for pdf, bitmapped graphics files
\usepackage{subfig}
\usepackage{epsfig} % for postscript graphics files
\usepackage{mathptmx} % assumes new font selection scheme installed
\usepackage{times} % assumes new font selection scheme installed
\usepackage{amsmath} % assumes amsmath package installed
\usepackage{amssymb}  % assumes amsmath package installed
\usepackage{color}
\usepackage{soul}
\usepackage[normalem]{ulem}
\usepackage{lipsum}

\newcommand{\comment}[1] {} %comment showed
\usepackage{amsmath} 
\usepackage{adjustbox}
\usepackage{multirow}
\usepackage{comment}
\usepackage[colorlinks,linkcolor=black]{hyperref}
\usepackage{threeparttable}
\usepackage{dsfont}

\usepackage{algorithm}
\usepackage[noend]{algpseudocode}

\usepackage{array}

\usepackage{parselines}

\title{\LARGE \bf Safe Trajectory Planning Using Reinforcement Learning for Self Driving }

\author{Josiah Coad$^{1}$, Zhiqian Qiao$^{2}$, John M. Dolan$^{2}$% <-this % stops a space
\thanks{$^{1}$Josiah Coad is with the Department of Computer Science \&
Engineering, Texas A\&M University,
College Station, TX, USA,
        {\tt\small josiahcoad@tamu.edu}}%
\thanks{$^{2}$Zhiqian Qiao and John M. Dolan\setlength {\marginparwidth }{2cm}
 are with The Robotics Institute, Carnegie Mellon University, Pittsburgh, PA, USA, 
        \{{\tt\small zhiqianq, john}\}{\tt\small @andrew.cmu.edu}}%
}

\begin{document}

\maketitle
\thispagestyle{empty}
\pagestyle{empty}

%%%%%%%%%%%%%%%%%%%%%%%%%%%%%%%%%%%%%%%%%%%%%%%%%%%%%%%%%%%%%%%%%%%%%%%%%%%%%%%%
\begin{abstract}
Self-driving vehicles must be able to act intelligently in diverse and difficult environments, marked by high-dimensional state spaces, a myriad of optimization objectives and complex behaviors. Traditionally, classical optimization and search techniques have been applied to the problem of self-driving; but they do not fully address operations in environments with high-dimensional states and complex behaviors. Recently, imitation learning has been proposed for the task of self-driving; but it is labor-intensive to obtain enough training data. Reinforcement learning has been proposed as a way to directly control the car, but this has safety and comfort concerns. We propose using model-free reinforcement learning for the trajectory planning stage of self-driving and show that this approach allows us to operate the car in a more safe, general and comfortable manner, required for the task of self driving.
  
\end{abstract}

%%%%%%%%%%%%%%%%%%%%%%%%%%%%%%%%%%%%%%%%%%%%%%%%%%%%%%%%%%%%%%%%%%%%%%%%%%%%%%%%
\section{Introduction}

Machine learning (ML) is an attractive approach to self driving, primarily because it is able to learn how to interact in a complex environment (too complex for rules) by minimizing a cost. Complexity comes from stochasticity, high-dimensional state spaces and/or intricate policies. Model-free reinforcement learning (RL) is a type of machine learning well-suited to self-driving, as it optimizes a policy which maximizes long-term rewards in a stochastic environment without the need for a model of the environment. Furthermore, RL does this without the need for laborious data collection, as in imitation learning. Furthermore, RL has been shown to outperform human experts on many games, so it could theoretically be better than human driver collected data (human error accounts for 90\% of traffic accidents, so training our ML policies on such data may not be optimal.)

RL equipped with deep learning (DL) is able to learn complex policies in high-dimensional state spaces and continuous action spaces. However, in using any type of DL for the self-driving task, the question arises of whether a DL policy can respect the many self-driving constraints that should never get breached, such as colliding with a pedestrian or exceeding some maximum centripetal acceleration that would cause the car to flip. It is nontrivial to impose such constraints on a DL policy which is often a black box system.

Many current state-of-the-art RL self-driving approaches, such as \cite{end-to-end-race, CIRL, end-to-end-deep, end-to-end-model, carla} use control-based RL that outputs the next actuator control (steer, throttle) for the vehicle. Most of these approaches train the policy end-to-end, meaning directly from camera input to car control output, thus bypassing the traditional pipeline of: perception, behavior choosing, motion planning, control. This has serious safety concerns since we have no idea how the DL model will steer and control the car until it generates the immediate next control.

\begin{figure}
\includegraphics[width=\columnwidth]{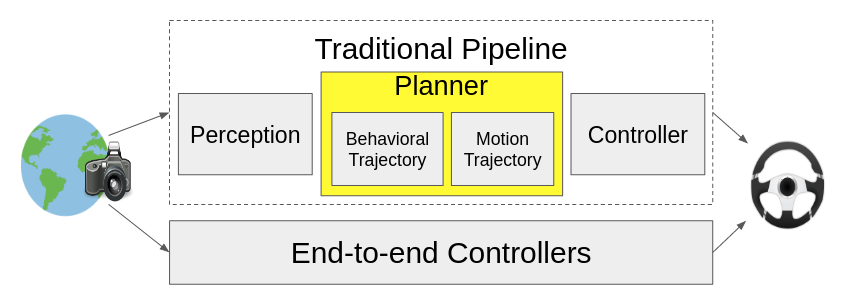}
\caption{A comparison of our method to end-to-end controllers. Hierarchical planners such as \textit{Focused} and \textit{Hierarchical} propose breaking planning into two parts. We propose using RL for generating the entire trajectory (shown in yellow).}
\centering
\label{fig:system_compare}
\vspace{-0.5cm}
\end{figure}

Instead of bypassing the entire pipeline, we propose keeping the modules distinct and using RL to generate the behavioral trajectory (seen in green in Figure \ref{fig:coordinates}). We assume a separate perception module to extract the occupancy grid and speed limits. We then use a RL-trained policy to plan a trajectory given that information. We finally use feed the trajectory to a proportional integral derivative (PID) controller to generate the vehicle controls. By using trajectory planning instead of direct control, we can apply higher level safety systems to restrict the RL-generated trajectory if it is seen to take the car into an unsafe state, such as off the road or into an obstacle. See Figure \ref{fig:system_compare} for a comparison of our method to the end-to-end control-based approach.

Our approach derives inspiration from several classical motion planning techniques, primarily \textit{Focused} \cite{focused-trajectory-planning-for} and \textit{Hierarchical} \cite{hierarchical-trajectory-planning-of}. These hybrid approaches seek to combine two previously used techniques: low-fidelity grid search and path optimization. In the first step, they perform an exhaustive search over all possible discrete paths to choose an optimal behavioral trajectory. In the second, they perform a non-convex numerical path optimization to produce the final trajectory. However, the authors of these papers note a problem with breaking this up into two steps: we could be constraining ourselves to a local minimum by choosing a behavioral trajectory is only optimal in the discrete space. Thus, we find motivation to directly produce a smoothed out trajectory, which we can do with RL. Additionally, using RL is comparatively computationally cheap at test time since it does not need to do an exhaustive search and numerical optimization at each step, but rather use a pretrained network to directly produce a trajectory.

In this paper, we propose a reinforcement learning based trajectory planner for self-driving. We show the advantage of this approach over both classical planning methods and reinforcement learning based controllers.

The structure of this paper is as follows: in Section \ref{sec:related-work} we present related works and how they motivate our approach. In Section \ref{sec:methods} we formalize the problem and our solution. In Section \ref{sec:experiments} we explain RL training details and our results. Finally, in Sections \ref{sec:conclusion} and \ref{sec:future-work} we summarize our contribution and provide directions for further work.

\section{Related Work}  \label{sec:related-work}

\subsection{Non-machine Learning Trajectory Planning}
% A trajectory is a sequence of local states, such as cartesian coordinates and velocity, parameterized by time, to be visited by the vehicle. Trajectory planning is the task of planning a feasible trajectory within some lookahead distance of the vehicle that does not result in a collision and moves towards a goal. An optimal trajectory should avoid obstacles such as other cars and pedestrians, adhere to a set of motion constraints and respect traffic rules \cite{real-time-motion-planning}. % Need this?

Many non-deep learning approaches to trajectory planning have been proposed, roughly fitting into the categories of sampling, search and numerical optimization. Some examples of sampling planners are rapidly expanding random trees \cite{motion-planning-for-car} and lattice sampling \cite{focused-trajectory-planning-for, local-Path-Planning-for, spatiotemporal-state-lattices-for}. Many of these systems make strong simplifying assumptions about the environment, such as no uncertainty in an obstacle's motions, or they severely discretize the planning space \cite{real-time-motion-planning}. These systems also require extensive human engineering to tune. Thus, many of the traditional trajectory planners are not able to handle complex environments marked by a large state space, multifaceted objectives and many obstacles. Numerical optimization methods such as seen in \cite{making-bertha-drive, trajectory-planning-for-bertha} also have been proposed for self-driving. In numerical optimization however, objects must be represented as differential mathematical functions, preferable convex, even though many states in the self-driving environment are discrete. This makes it difficult under the numerical optimization approach to consider all the objects that exist in a dense urban environment. 

Our approach derives inspiration from several non-ML-based hierarchical methods. A hierarchical approach to trajectory planning was first proposed in \textit{Focused} \cite{focused-trajectory-planning-for}. The two-step hierarchy starts with generating a low-fidelity spatiotemporal path through a node-layer network superimposed on the road. The authors use exhaustive search over all possible discrete paths to choose an optimal path through the road network. In the second step of this approach, the chosen path is fine-tuned. \textit{Focused} was tested in single lane roads with sparse static obstacles.

%\textit{Focused} does not address planning with dynamic obstacles or in the presence of regulator road elements (e.g. speed bumps). Thus, \textit{Focused} is limited to roads without regulator elements where the ego vehicle is the only moving agent in the environment.

\textit{Hierarchical} \cite{hierarchical-trajectory-planning-of} builds on \textit{Focused} by also accounting for regulator elements such as stoplights and speed bumps. They first generate a low-fidelity path which they call a behavioral trajectory because the path and velocity profile chosen are greatly influenced by environmental factors such as obstacles, regulatory elements and road geometry and thus require behavioral choices. They then use this to create a fine-tuned trajectory called the motion trajectory.

%We build off this approach but replace the behavioral trajectory generation module with a RL-learned policy. Figure \ref{fig:system_compare} shows the module we replace with RL, and Figure \ref{fig:behave} zooms in on this two step trajectory planner.

In the past, behavioral choice has been implemented with finite state machines \cite{road-model-based-and, junior-the-stanford-entry, development-of-autonomous-car}. However, as \textit{Hierarchical} points out, there are issues with this such as defining an interface between the behavior choice and the trajectory planner. Thus, they point out that subsuming the behavioral planning into the trajectory planning can result in a more streamlined approach. RL has been proposed for the behavioral module previously, so we now propose to use RL for the behavioral trajectory generation.

For a more comprehensive review of non-machine learning approaches to trajectory planning and their limitations, refer to the surveys \cite{real-time-motion-planning, a-survey-of-motion, a-review-of-motion}.

\subsection{Imitation Learning-Based Trajectory Planning} \label{related-work-imitation-learning}

Currently, imitation learning is the predominant deep learning-based approach used for trajectory planning. A straightforward way taken by \cite{deep-imitation-learning-for} and others is to train a neural network to take camera input and output a layer of size $2H$ where $H$ is the size of the horizon. The output represents a trajectory path $\left<x_{t+1}, y_{t+1}, \dots , x_{t+H}, y_{t+H}\right>$. The goal is to minimize the displacement $d_{t+i}$ between the expert’s actual trajectory point $(x_{t+i}, y_{t+i})$ and the predicted point $(\hat{x}_{t+i}, \hat{y}_{t+i})$:
$$d_{t+i} = ((x_{t+i} - \hat{x}_{t+i})^2 + (y_{t+i} - \hat{y}_{t+i})^2)^{\frac{1}{2}}$$
The loss is the mean squared error of the displacements. ChauffeurNet \cite{chauffeurnet} does this with the addition of perturbations when collecting the trajectory samples for the sake of robustness. 

In \cite{deep-imitative-models-for}, the authors do something slightly different by aiming to learn the distribution parameters which maximize the likelihood of an expert path conditioned on the agent's observation. They fit the imitative model in a supervised manner using a recurrent neural network. In a subsequent paper, NeuroTrajectory \cite{neurotrajectory}, the authors use imitation learning and evolutionary methods to learn to imitate expert drivers. Learning by Cheating \cite{learning-by-cheating} learned to extract a high-level state space and then used that to train an imitation learning-based planner.

Imitation learning still faces limitations, however, in that it requires large amounts of human driver training data, which is laborious to collect. Furthermore, the learned policy tends to overfit to the environment which it was trained on and could fail if presented with a vastly different situation, such as recovering from a near-crash scenario, if this scenario was not sufficiently covered in the training data. Thus, the trained algorithm may not react appropriately in these conditions in deployment. A recent paper argues that even with tens of millions of examples, direct imitation learning sometimes does not yield satisfactory driving policies \cite{chauffeurnet}.

\subsection{End-to-End Reinforcement Learning Controllers}

It is currently a common approach to use reinforcement learning to learn an end-to-end policy for directly controlling the car. In End-to-End Race Driving \cite{end-to-end-race}, the authors train an RL agent to drive a race car around a track by directly controlling the car with the agent. However, the ego vehicle exhibits aggressive unnecessary steering behavior and is not tested in any environment that includes other vehicles. 

In \cite{driving-in-dense-traffic}, the authors use an RL agent to control the derivative of acceleration and steer, which is jerk and steering rate. They claim this increases the comfort of the ride by reducing oscillatory and jerky behavior.

In the release of the CARLA simulator \cite{carla}, the authors introduce a baseline end-to-end RL system that they used to compare in urban driving performance to a modular pipeline approach and an end-to-end IL approach. In their results, they found the end-to-end RL control approach to have inferior performance to the other approaches. 

In the recent work \cite{end-to-end-model}, the authors note how difficult it is to get a good policy directly from a front camera. In response, they separately train the perception module to derive high-level features about the world and then train a control module, given the extracted world information. In CIRL \cite{CIRL} and \cite{end-to-end-model} the authors introduce a gating unit which is used to choose which high-level behavior to follow (lane-following, left, right or straight) in urban driving. However, even in their best results, all these works still report crashes, motivating the need for future work. 

Additional work in this area can be found in the surveys \cite{a-survey-of-deep, a-survey-of-end}. The end-to-end control approach, while attractive for simplicity of implementation (the idea that one black box does it all), lacks interpretability and modularity, making it dubious for real world application in self-driving. Specifically, this approach is difficult to debug and to add higher-level logic to the system. Furthermore, control-based approaches have known problems of being jerky/oscillatory—see demo\footnote{\url{https://youtu.be/e9jk-lBWFlw}} from \cite{end-to-end-race}. In addition, controls must be generated at a high and consistent frequency; and thus the system is brittle to any unexpected system latency. Finally, even in low-speed conditions when tested in simulators, these control-based approaches still suffer collisions, showing that more safety guarantees are needed before an RL-based solution can be used in industry.

Applying a safety constraint to a RL generated control, i.e. revoking the next steer or throttle if it is deemed that such an action would take the car into an unsafe state, would not work, since the car is often travelling at speeds that cannot be instantaneously stopped or diverted. Our approach of using RL for trajectory planning seeks to remedy these issues. Specifically, by planning a trajectory instead of directly controlling the car, we create a more interpretable and modular system and take advantage of that fact by applying a high-level safety constraint on the RL action to ensure safety of the action before execution.

\section{Methodology} \label{sec:methods}

\subsection{Trajectory Planning}

The goal of trajectory planning is to find a sequence of states $\vec{q}=\left<q_1, q_2, ...q_n\right>$ where $q_i=(x_i, y_i, speed_i)$ indicates the path that the ego vehicle should travel to avoid obstacles, respect road rules and maximize comfort and efficiency towards a goal. Once we obtain $\vec{q}$, we can pass it on to a controller, such as a PID controller. The PID outputs controls $c$ to track the path where $c=(steering, throttle, brake)$ which is applied directly to the actuators of the car.

\subsection{Reinforcement Learning}

Reinforcement learning seeks to learn an optimal policy $\pi$  parameterized by $\theta$ which maps a state (observation) $s$ in the observation space $\mathbb{S}$ to an action $a$ in the action space $\mathbb{A}$, i.e. $\pi_{\theta}(a|s)$. In deep RL, $\theta$ are the weights to a neural network. Unlike supervised learning, RL does not require any labelled training data. Instead, RL interacts with an environment by being presented with $s_t$ and choosing $a_t$ according to its policy $\pi$. Under the assumption of full observability, RL operates on the premise that the environment can be viewed as a Markov decision process with a (possibly stochastic) transition function $T(s_{t+1}|s_t, a_t)$. The environment also returns a reward $R(s_t, a_t)$. The process of observing $s_t$, getting action $a_t$ according to the policy $\pi$, and executing $a_t$ in the environment is called a step. The reinforcement algorithm works to find the policy parameterization that maximizes the expected reward over the episode:
\begin{equation} \label{eq:rl}
argmax_{\theta} \mathbb{E} [R(s_t,a_t) | a_t \sim \pi_{\theta}(s_t), s_t \sim T(s_{t-1}, a_{t-1})]
\end{equation}

Both continuous and discrete actions are possible with reinforcement learning, depending on the RL algorithm used. In the case of discrete, $\mathbb{A}$ is a (generally small) finite action space. In the case of continuous, $\mathbb{A}$ has a shape of $[-1, 1]^n$ where $n$ is the number of controls contained in one action. The constraint to $[-1, 1]$ is trivial since we can scale the action once inside the step function. However, empirically this constraint is better for convergence. In our case, $n=2H$ where $H$ is the number of layers we are planning out to and 2 because for each point we pick, we also pick an associated velocity.

We use a reinforcement learning algorithm called Proximal Policy Optimization (PPO) \cite{ppo}. PPO is a variant of the actor critic family that also includes an entropy term in its loss function to encourage exploration and clips policy changes so that policies do not update too drastically. It should be noted that our approach is not limited to use with PPO.

\subsection{Safety Constraint}

Because we are planning seconds into the future, we can detect potential traffic violations, kinematic violations and collisions with obstacles far before they happen and account for them. Call the planning space that results in one of these undesirable outcomes $\neg C^{free}$. Being able to account for collisions and violations is an advantage over reactive control-based RL, in which we don't know what control the RL agent is going to choose next until it is too late.

We need to somehow constrain the RL-chosen action to free space during planning, notated as $C^{free}$, which considers the occupancy grid and maximum centripetal acceleration. To do this, we first generate an action/trajectory proposal from our RL policy, calling this $a$. We then take the $\ell^2$ norm of the residuals from a projection of $a$ to $C^{free}$ space and call the element in $C^{free}$ that minimizes the $\ell^2$ norm of the residuals, $u^*$, as seen in Equation \ref{eq:projection}.

\begin{equation} \label{eq:projection}
u^* := P_{C^{free}}(a) = argmin_{u\in C^{free}}||a-u||_2^2
\end{equation}

If the greatest absolute distance between the proposed trajectory and the projection is within some tolerance $\tau$ (we use $\tau=0.5$), we allow the proposed action as is, else we choose $u^*$. If the Safety Constraint determines that there is no safe space, it throws an exception that can be handled by an emergency stopping module. This procedure is formalized in Algorithm \ref{algo:constraint}.

\begin{algorithm}
\caption{Constraining RL Action to Safe Space}
\label{algo:constraint}
\begin{algorithmic}
\Function{Safety-Constraint}{proposed trajectory $a$}
    \State find $C^{free}$
    \If {$C^{free}$ is empty}
        \State THROW $NoPathException$
    \EndIf
    \State $u^* \gets P_{C^{free}}(a)$ 
    \If {max $|u^*-a| < \tau$}
        \State \Return a
    \Else
        \State \Return $u^*$
    \EndIf
\EndFunction
\end{algorithmic}
\end{algorithm}

Using Algorithm \ref{algo:constraint}, we can offer a much stronger assurance than control-based RL that collisions and other violations will not be encountered during deployment.

By allowing some tolerance in the path, we are allowing the RL to potentially chose a better path than what was in the discretized options. Yet, by restraining the deviance of the path, we still make a strong case for safety within the limits of our perception.

\subsection{Environment}

\begin{figure}
\includegraphics[width=0.48\textwidth]{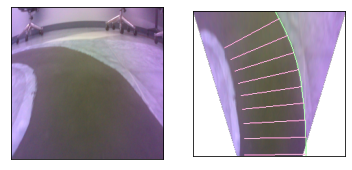}
\caption{Left is the view of the road from a front camera. Right is the result of the perception processing pipeline: transform to bird's-eye via a homography, run semantic segmentation to identify the road, fit third order polynomials to the road bounds and finally, derive normal lines to the edges to get the layers which we will plan along.}
\centering
\label{fig:line_detect}
\end{figure}

\begin{figure}
\includegraphics[width=0.48\textwidth]{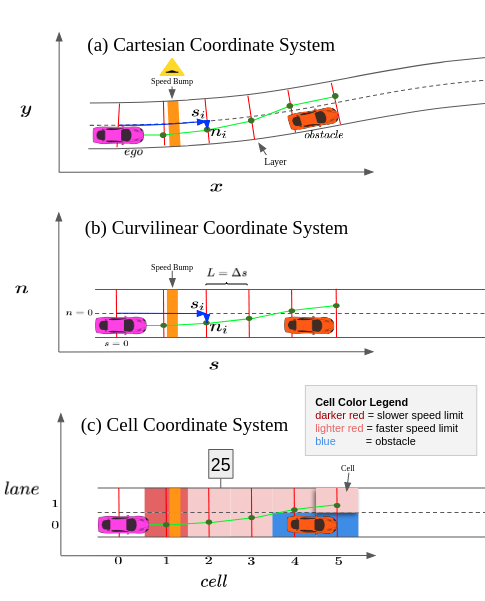}
\caption{Mapping a bird's-eye view (a) to a cell coordinate system (c) in which we plan. $(s_i, n_i)$ shown in (a) gets mapped to the respective location in (b). Distance between layers is a function of speed and is denoted $L$. $H$ is the sensor distance, i.e. how far we plan out to.}
\centering
\label{fig:coordinates}
\end{figure}

Once we obtain a bird's-eye view of the scene, we can map the Cartesian coordinate system to a curvilinear space to aid in planning. To do this, we first fit a curve to the road. In this paper, we use a third degree polynomial; but other curves such as a spline can been used. Next, equidistant perpendiculars can be drawn to this curve that make up layers along the road. The distance between layers is a function of speed. See Figure \ref{fig:line_detect} for a demonstration of the transformation to bird's-eye. Now we have the scene in Cartesian (real-world) coordinates as seen in Figure \ref{fig:coordinates}(a). The distance along this curve, zeroed at the ego vehicle, makes the s curvilinear axis. Distance along the perpendiculars makes our n axis, as seen in Figure \ref{fig:coordinates}(b). A point $(x_i, y_i)$ can be mapped to $(s_i, n_i)$ by using a closest-point method such as quadratic minimization \cite{robust-and-efficient-computation}. A point $(s_i, n_i)$ can be mapped to $(x_i, y_i)$ by moving along the original curve $s_i$ distance and then moving $n_i$ distance along the perpendicular. Finally, we convert curvilinear to a cell coordinate system by simply dividing $n$ by lane width and $s$ by layer distance ($L$), as seen in Figure \ref{fig:coordinates}(c).

\subsection{State Space}

As outlined in the previous subsection, we obtain a cellular representation of the environment. Our state space $\mathbb{S}$ includes the static occupancy grid/matrix $O \in \{0, 1\}^{H\times W}$ and the speed regulatory grid $S \in \mathbb{R}^{H\times W}$ of those cells. $H$ is the cellular depth of our sensor range, i.e. how far we plan out to and $W$ is the cellular width of the road. The static occupancy grid can be reasonably extracted from LIDAR. The speed regulatory grid can be extracted via an identification of regulatory elements (speed limit sign, stop sign, speed bump, etc.) using a camera, as was done in \textit{Hierarchical}. Finally, the current lateral position $n_0$ and speed $v_0$ are included in the state space. In summary, $\mathbb{S}:=\{O,S,n_0,v_0\}$. $O$ and $S$ can be rendered graphically as seen in Figure \ref{fig:coordinates}(c). Blue represents a 1 in $O$, i.e. the space is occupied. Shades of red represent entries in $S$ with darker red representing slower speed limits. 

Currently, $\mathbb{S}$ is used as both a way to check if the proposed action is in $C_{free}$ and as an input to our policy. However this is not a requirement; and indeed, it is an interesting direction for future work to feed a more complex state space to the policy, such as front camera footage, and only use $\mathbb{S}$ for the subsequent safety constraint.

\subsection{Action Space}

The action space for the RL agent is of size $2H$ where $H$ is the number of layers we are planning through, spaced out by some distance $L$ between the layers in curvilinear space. The action space is defined as $a:=\{[\Delta n|\Delta v] \in [-1,1]^{2H}$ where $\Delta p_j$ is the lateral change between layer $j-1$ and $j$ and $\Delta v_j$ is the velocity change between layer $j-1$ and $j$ (where $j=0$ is defined as the current state). This action space can be converted to points $n_j$ along each layer $l_j$ by taking the cumulative sum, i.e. $n_j=\sum_{i=0}^{j}{\Delta p_j}$. Similarly, the same can be done for each respective $v_j$ at each $n_j$. 

\subsection{Reward Structure}

The reward function is made of a trajectory cost $f$ which is a linear combination of chosen weights and the following terms: $f^r$, $f^a$, $f^j$, $f^d$, $f^k$, $f^l$, $f^c$. Each term is calculated by summing the individual components along the trajectory, i.e. $f^*=\sum_{i=0}^H f^*_i$. The $f$ terms are described in Table \ref{tab:reward-settings}. Additionally, a success is given a reward of 10, a failure is given a reward of -20 and each step is given a reward of 1. 

\begin{table}
\begin{center}
\caption{Reward Structure Terms}
\begin{tabular}{ | c | c | c |}
\hline
\textbf{Cost Term} & \textbf{Name} & \textbf{Formula}\\ 
\hline
$f^r_i$ & speed error & $(v_{i}^{ref}-v_{i})^2$ \\
\hline
$f^a_i$ & acceleration & $(v_{i}-v_{i-1})^2 / (2\cdot dist_i)$ \\
\hline
$f^j_i$ & jerk & $f^a_{i+1} - f^a_i$ \\
\hline
$f^d_i$ & extra distance & $dist_i - L$ \\
\hline
$f^k_i$ & curvature & $(n_{i+1}-2n_i+n_{i-2}) / L$ \\
\hline
$f^l_i$ & lane crossing & $\mathds{1}$[$n_i$, $n_{i-1}$ in different lanes] \\
\hline
$f^c_i$ & centripetal acceleration & $f^k_i \cdot v_i$ \\
\hline
\end{tabular} \\
\label{tab:reward-settings}
\end{center}
\end{table}

\section{Experiments} \label{sec:experiments}

\subsection{Training details}

\begin{figure}
\includegraphics[width=0.48\textwidth]{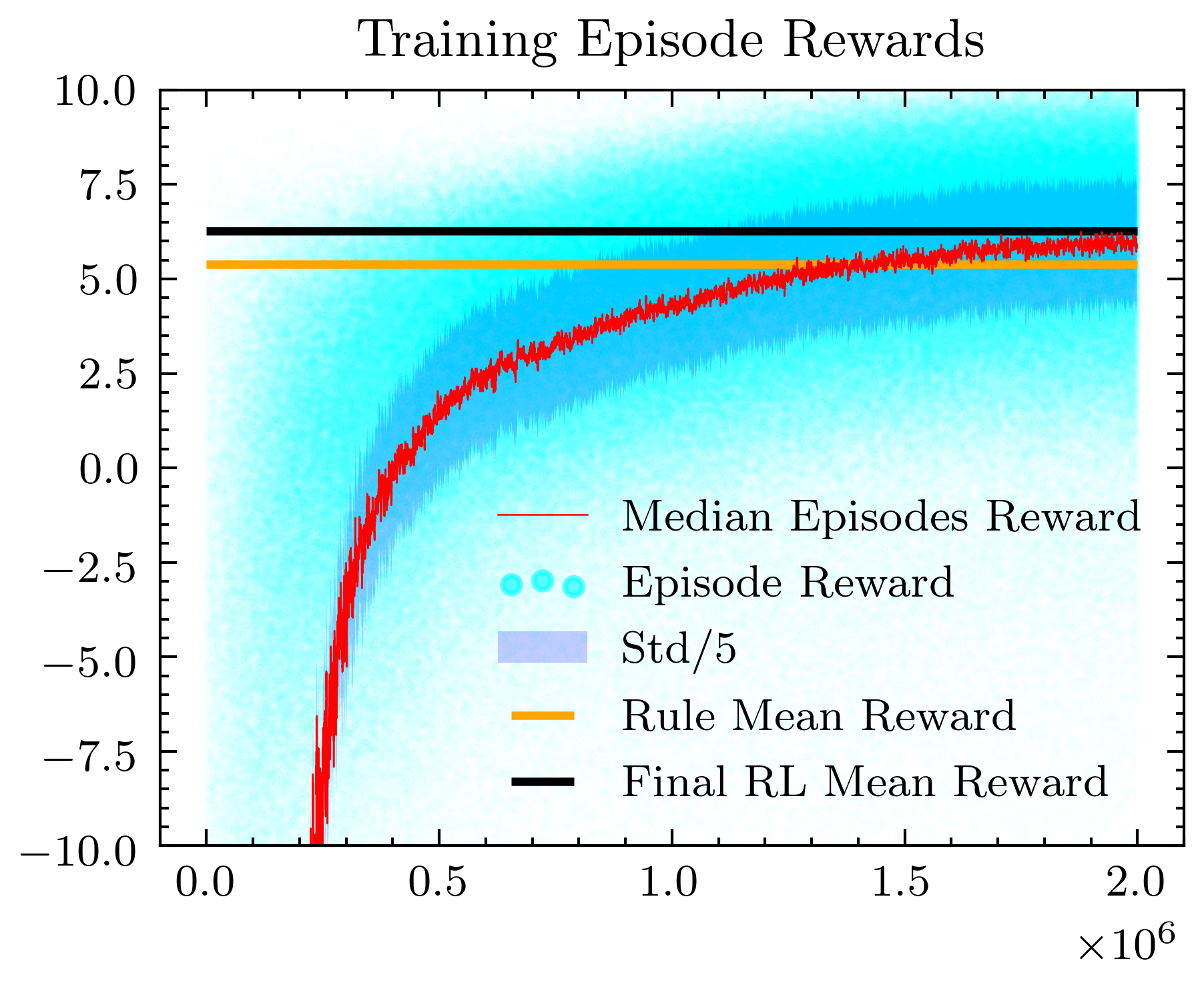}
\caption{Average episode rewards from two million steps of training for the RL trajectory planner. Red line is rolling window median with 1000 window length.}
\centering
\label{fig:training-rewards}
\end{figure}

In Figure \ref{fig:training-rewards} we show some of the episode rewards during training from 2M steps. The training was run on a Tesla K80 GPU with 32 parallelized environments using the PPO RL algorithm and took about 4 hours of wall time in our custom environment.

Each training episode was initialized with static obstacles appearing uniformly randomly over each cell drawn from a Bernoulli$(p=0.5)$ distribution. During training, we ensured that randomly generated obstacles did not fully block the road, until max-steps was reached. Once max-steps was reached, we blocked off the road with a wall. During training, the car planned 3 layers ahead, moved 3 layers and then replaned. During testing, the car planned 3 ahead and moved 1 layer before replanning.

\subsection{Network Architecture}
To represent the policy, we used a fully-connected (FC) multilayer perception network with two hidden layers of 64 units and tanh nonlinearities. The policy outputs the mean of a Gaussian distribution with variable standard deviations. We don’t share parameters between the policy and value function (so coefficient $c_1$ from the original PPO paper is irrelevant.)

In Table \ref{tab:ppo-hyperparams} we outline the hyperparameter settings we found to work best. We used the Python package called Stable Baselines (SB) version 2.1 \cite{stable-baselines} for their implementation of Proximal Policy Optimization (PPO) called PPO2.

\begin{table}
\begin{center}
\caption{Hyperparmeter settings for PPO}
\label{tab:ppo-hyperparams}
\begin{threeparttable}[b]
\begin{tabular}{ | m{10em} | c | c | }
\hline
\textbf{PPO Paper Name} & \textbf{SB PPO2 Name} & \textbf{Value} \\ 
\hline
Number of actors & n\_envs & 32 \\
\hline
Horizon (T) & n\_steps & 64 \\
\hline
Minibatch size\tnote{1} & nminibatches & 64 \\
\hline
Discount ($\gamma$) & gamma & 0.999 \\
\hline
Adam stepsize & learning\_rate & $2e^{-4}$ \\
\hline
Entropy coefficient ($c_2$)& ent\_coef & $0.01$ \\
\hline
Clipping epsilon & cliprange & 0.4 \\
\hline
Number of epochs & noptepochs & 25 \\
\hline
GAE parameter ($\lambda$) & lam & 0.99 \\
\hline
Policy network & net\_arch vf & FC: [64, 64] \\
\hline
Value network & net\_arch pi & FC: [64, 64] \\
\hline
\end{tabular}
\begin{tablenotes}
\item [1] nminibatches was set to 64 in the SB PPO2 implementation. However, this equals a minibatch size of 32 since minibatch\_size = (n\_envs*n\_steps)/nminibatches.
\end{tablenotes}
\end{threeparttable}
\end{center}
\end{table}

\subsection{Results}

% Next, we compare our method to control-based RL. We show that our method produces a much smoother and safer ride. 

% \textbf{TODO: Compare with control based.}

% \textbf{TODO: Show constraint mechanism.}

% \textbf{TODO: Run N trials and calculate cost metrics across}

\begin{figure}
\includegraphics[width=0.48\textwidth]{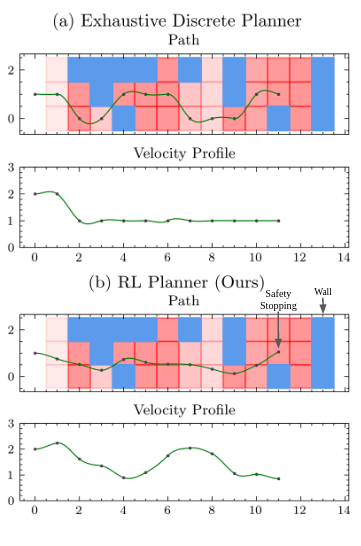}
\caption{In the Path figure, a three-lane road randomly populated with static obstacles and regulatory speeds is shown in cellular units. A wall appears at the end of the road. The purple dots represent points along each layer that the agent choose for the trajectory. In (a), the discrete exhaustive search chooses the optimal trajectory. In (b), we use RL to choose the optimal trajectory in a continuous action space along each layer. Green is a cubic spline which has been fit to the purple dots to form a path to travel along. In the Velocity Profile figure, the respective speeds chosen for each layer are shown. Numerical analysis of the paths are presented in Table \ref{tab:results-compare-stats}.}
\centering
\label{fig:results-compare}
\end{figure}

In Figure \ref{fig:results-compare}, we can see one of the advantages of using RL that plans in continuous search space. We compare to the discretized exhaustive search method and see that we obtain a smoother behavioral path with less drastic turns. This is also shown in Table \ref{tab:results-compare-stats} where we see that the RL results in a smoother and more efficient path.

Furthermore, using RL results in less latency to query for the next trajectory. One exhaustive search in the relatively small action space we used takes an average of $7.9e^{-3}$ seconds, while querying for the next trajectory from the RL policy takes $8.7e^{-4}$ seconds, making RL nearly a magnitude faster. This speed difference is crucial in real time computing requirements such as self-driving. Finally, we can see how the car comes to a safety stop before colliding with the wall as a result of the Safety Constraint from Algorithm \ref{algo:constraint}. In static environments, assuming enough plan-ahead distance and proper perception, we can ensure 100\% success rate, i.e. no collisions.

\begin{table}
\begin{center}
\caption{Driving Metrics from Figure \ref{fig:results-compare}}
\begin{tabular}{ | c | c | c | c |}
\hline
\textbf{Measure} & \textbf{Aggregation} & \textbf{Exhaustive} & \textbf{RL (Ours)}\\ 
\hline
Step Reward & mean & -1.49 & \textbf{-1.14} \\
\hline
Acceleration & max & 1.0 & 1.0 \\
\hline
Jerk & max & 1.4 & \textbf{0.3} \\
\hline
Extra Distance & mean & 0.3 & \textbf{0.1} \\
\hline
Curvature & max & 1.3 & \textbf{0.8} \\
\hline
Centrip. Acc. & max & 2.0 & \textbf{1.3} \\
\hline
\end{tabular} \\
\label{tab:results-compare-stats}
\end{center}
\end{table}

\begin{table}
\begin{center}
\caption{Driving Metrics over 1000 Episodes}
\begin{tabular}{ | c | c | c | c |}
\hline
\textbf{Measure} & \textbf{Aggregation} & \textbf{Exhaustive} & \textbf{RL (Ours)}\\ 
\hline
Speed Track Err & Mean & 7.582 $\pm$ 0.195 &  \textbf{5.306} $\pm$  0.126 \\
\hline
Acceleration & Max  & 0.596 $\pm$ 0.017 &  \textbf{0.367} $\pm$  0.012 \\
\hline
Jerk & Max  & 0.407 $\pm$ 0.012 &  \textbf{0.194} $\pm$  0.005 \\
\hline
Excess Distance & Mean & 0.317 $\pm$ 0.011 &  \textbf{0.185} $\pm$  0.007 \\
\hline
Curvature & Max  & 0.738 $\pm$ 0.026 &  \textbf{0.304} $\pm$  0.010 \\
\hline
Lane Changes & Sum  & \textbf{0.791} $\pm$ 0.027 &  0.960 $\pm$  0.029 \\
\hline
Centrip. Acc. & Max  & 1.367 $\pm$ 0.058 &  \textbf{1.147} $\pm$  0.036 \\
\hline
\end{tabular} \\
\label{tab:results-compare-stats-1000}
\end{center}
\end{table}

\section{Conclusion} \label{sec:conclusion}
In this work, we are the first to our knowledge to propose using model-free reinforcement learning for the task of trajectory planning in self-driving. By using this approach, we garner the benefits of looking ahead by being able to apply safety constraints to our actions and we take advantage of the continuous action space of the RL agent. We show how our method has an advantage over control-based RL in that it results in less collisions and smoother rides. Furthermore, we show that it has an advantage over exhaustive discrete search-based methods in that it is computationally faster and results in smoother trajectories. The safety, interpretability, comfort and speed advantages of our system aim to make RL a viable option for future deployment in industry.

\section{Future Work} \label{sec:future-work}
We are interested in extending this system to use inverse reinforcement learning to learn the reward function weights (instead of a human engineer picking all the tuning parameters as is now required). We also believe that incorporating an LSTM into our approach would allow capturing the partially observable (POMDP) nature of driving and would generalize our approach. We also wish to extend our approach to incorporate front camera data directly into our policy via a CNN, which would better take full advantage of the high capacity of our machine learning approach. We would also like to plan in dynamic environments and implement this system in a real world car.

\section{Acknowledgements}
This material is based upon work supported by the National Science Foundation under Grant No. 1659774. We would like to thank the organizers of the Carnegie Mellon Robotics Institute Summer Scholars (RISS) program for making this experience possible.

% \addtolength{\textheight}{-8cm}   % This command serves to balance the column lengths

%%%%%%%%%%%%%%%%%%%%%%%%%%%%%%%%%%%%%%%%%%%%%%%%%%%%%%%%%%%%%%%%%%%%%%%%%%%%%%%%

%%%%%%%%%%%%%%%%%%%%%%%%%%%%%%%%%%%%%%%%%%%%%%%%%%%%%%%%%%%%%%%%%%%%%%%%%%%%%%%%

%%%%%%%%%%%%%%%%%%%%%%%%%%%%%%%%%%%%%%%%%%%%%%%%%%%%%%%%%%%%%%%%%%%%%%%%%%%%%%%%

%%%%%%%%%%%%%%%%%%%%%%%%%%%%%%%%%%%%%%%%%%%%%%%%%%%%%%%%%%%%%%%%%%%%%%%%%%%%%%%%

\bibliography{behavior}
\bibliographystyle{IEEEtran}

\end{document}